\documentclass[10pt,twocolumn,letterpaper]{article}

\usepackage{cvpr}
\usepackage{times}
\usepackage{epsfig}
\usepackage{graphicx}
\usepackage{amsmath}
\usepackage{amssymb}

\usepackage{xcolor}
\usepackage[T1]{fontenc}
\usepackage{adjustbox}
\usepackage{comment}

\usepackage{bm}
\usepackage{pifont}
\newcommand{\cmark}{\ding{51}}%
\newcommand{\xmark}{\ding{55}}%

\newcommand{\red}{\textcolor{black}}
\newcommand{\blue}{}

\usepackage{lipsum}

\newcommand\blfootnote[1]{%
  \begingroup
  \renewcommand\thefootnote{}\footnote{#1}%
  \addtocounter{footnote}{-1}%
  \endgroup
}

\usepackage[pagebackref=true,breaklinks=true,letterpaper=true,colorlinks,bookmarks=false]{hyperref}

\DeclareMathOperator*{\argmin}{argmin}
\cvprfinalcopy 

\ifcvprfinal\fi
\begin{document}

\title{FSS-1000: A 1000-Class Dataset for Few-Shot Segmentation}

\author{Xiang Li$^{1}$\hspace{0.19in} Tianhan Wei$^1$\hspace{0.19in}  Yau Pun Chen$^{1}$\hspace{0.19in} Yu-Wing Tai$^2$\hspace{0.19in} Chi-Keung Tang$^1$\\
$^1$HKUST \hspace{1.5in}
$^2$Tencent\\
{\tt\small \{xlide, tweiab, ypchen\}@connect.ust.hk, yuwingtai@tencent.com, cktang@cs.ust.hk}}

\maketitle

\begin{abstract}
Over the past few years, we have witnessed the success of deep learning in
image recognition thanks to the availability of large-scale
human-annotated datasets such as PASCAL VOC, ImageNet, and COCO.
Although these datasets have covered a wide range of object categories,
there are still a significant number of objects that are not included.  Can
we perform the same task without a lot of human annotations? In this
paper, we are interested in few-shot object segmentation where the number
of annotated training examples are limited to 5 only. To evaluate and
validate the performance of our approach, we have built a few-shot
segmentation dataset, FSS-1000, which consists of 1000 object classes with
pixelwise annotation of ground-truth segmentation. Unique in FSS-1000, our
dataset contains significant number of objects that have never been seen
or annotated in previous datasets, such as tiny daily objects,
merchandise, cartoon characters, logos, etc.

We build our baseline model using standard backbone networks such as
VGG-16, ResNet-101, and Inception.  To our surprise, we found that
training our model from scratch using FSS-1000 achieves comparable and even better results than training with weights pre-trained by ImageNet which is more
than 100 times larger than FSS-1000. Both our approach and dataset are
simple, effective, and easily extensible to learn segmentation of new
object classes given very few annotated training examples. \red{Dataset is available at \url{https://github.com/HKUSTCV/FSS-1000}}
\blfootnote{\red{This research is supported in part by Tencent and the Research Grant
Council of the Hong Kong SAR under grant no. 1620818.}}

\end{abstract}

\section{Introduction}
Although unprecedented in the number of object categories when first released,
contemporary image datasets for training deep neural networks such as
PASCAL VOC~\cite{voc} (19,740 images, 20 classes), ILSVRC~\cite{ilsvrc} (1,281,167 images,
1,000 classes),  and COCO~\cite{coco} (204,721 images, 80 classes) are actually quite
limited for visual recognition tasks in the real world: a rough estimate of the
number of different objects on the Earth falls in the range of 500,000 to
700,000, following the total number of nouns in the English language.
While the exact total number of visual object categories is smaller than these
numbers, these large-scale datasets contribute less than 1\% in total.
Extending a new object category to existing datasets is a major
undertaking because a lot of human annotation effort is required: in
ImageNet, the mean number of images in a given class is 650.  More importantly, observe
that the number of images within each object category in ImageNet for
instance can vary significantly, ranging from 1 to 3,047.  This inevitably introduces undesirable biases which may have
a detrimental effect on important tasks solely relying on pre-trained weights obtained using a dataset that is biased in
both the choice of object classes (small number) and images within a given
class (uneven distribution). Biases in existing datasets have also been recently reported~\cite{bias_iclr19, focalloss}.

Thus, Few-Shot Learning has emerged as an attractive alternative for important computer vision tasks, especially when the given new dataset is very small and dissimilar so relying on the aforementioned pre-trained weights may not work well. Particularly relevant is image segmentation which requires extremely labor-intensive, pixelwise labeling for supervised learning. In few-shot segmentation, given an input consisting of a small support image set with labels (5 in this paper) and a query image set without labels, the learned model should properly segment the query images, even the pertinent objects belong to an object class unseen before.

There is {\em no} large-scale object dataset for few-shot segmentation. Previous research on few-shot segmentation relies on a manual split of the PASCAL VOC dataset to train and evaluate a new model~\cite{oslsm, iclr18}, but only 20 and 80 classes in the PASCAL VOC and COCO datasets respectively contain pixelwise segmentation information. Thus, building a large-scale object segmentation dataset is necessary to extensively and objectively evaluate the performance of our and future few-shot models.

FSS-1000 is the first large-scale dataset for few-shot segmentation with built-in object category hierarchy which emphasizes the number of object {\em classes} rather than the number of images.  FSS-1000 is highly scalable: 10 new images with ground-truth segmentation are all it takes for new object class extension.

\begin{table*}
\vspace{-0.1in}
\begin{center}
\begin{adjustbox}{max width=0.8\linewidth}
\begin{tabular}{c|ccccccc}
\hline
Dataset & Images & Classes & Classification & Detection & Segmentation & Mean & Stddev\\
\hline
SUN~\cite{sun} & 131,067 & 3,819 & \cmark  & \cmark & \xmark & 39.22 & 717.68 \\
ImageNet & 3,200,000 & 5,247 & \cmark  & \cmark & \xmark & 650.02 & 526.03 \\
Open Image & 9,052,839 & 7,186 & \cmark & \cmark & \xmark & 1409.62 & 14429.29 \\
PASCAL VOC 2012 & 19,740 & 20 & \cmark & \cmark & \cmark & 215.90 & 164.07 \\
MS COCO & 204,721 & 80 & \cmark & \cmark & \cmark & 4492.13 & 7487.38 \\
FSS-1000 & 10,000 & 1,000 & \cmark & \cmark & \cmark & 10 & \textbf{0} \\
\hline
\end{tabular}
\end{adjustbox}
\end{center}
\caption{Large-scale datasets comparison. Mean and standard deviation are
based on the expected number of images in each class.}
\label{tab:compare}
\vspace{-0.2in}
\end{table*}

\begin{figure}[t]
\begin{center}
   \includegraphics[width=1.1\linewidth]{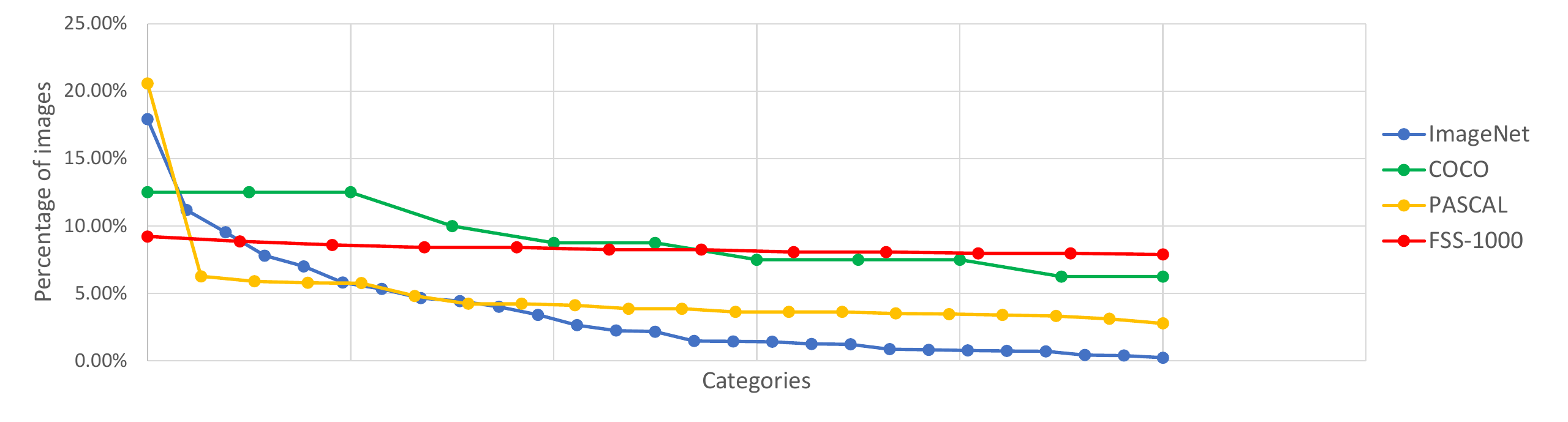}
\end{center}
\vspace{-0.1in}
   \caption{Normalized image distribution.  To make these datasets
comparable, we normalize each dataset respectively in the total number of
images ($y$-axis) and in the total number of object super-categories
($x$-axis) such that the area under each curve is 1 to make them
comparable.  All existing datasets are biased toward a number of object
categories except FSS-1000 (red).}
\label{fig:compare}
\vspace{-0.2in}
\end{figure}

Our baseline network architecture is constructed by appending a decoder module to the relation network~\cite{ltc}, which is a simple and elegant deep model effective and originally designed for few-shot image classification only. Reshaping the relation network into a fully-convolutional U-Net architecture~\cite{unet}, our extensive experimental results show that this baseline model trained from scratch on FSS-1000, which is less than 1\% of the size of contemporary large-scale datasets, outperforms the model fine-tuned from weights pre-trained on ImageNet/COCO dataset. In addition, without any fine-tuning / re-training, our trained baseline network can be applied to any unseen classes directly with decent performance. With its excellent segmentation performance as well as extensibility, FSS-1000 and our baseline model are expected to make a lasting contribution to few-shot image segmentation.
Please also refer to the supplemental materials for our extensive experimental results.

\begin{figure*}[t]
\begin{center}
   \includegraphics[width=1.0\linewidth]{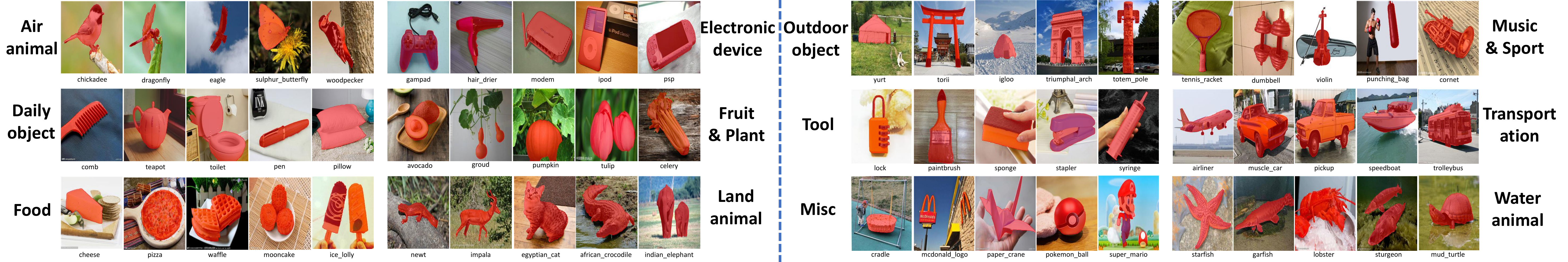}
\end{center}
\vspace{-0.1in}
   \caption{Example images and their corresponding segmentation in FSS-1000.
For the 12 super-categories here, 5 examples are shown, where the
ground-truth segmentation map is overlaid in red in the corresponding
image.}
\label{fig:examples}
\vspace{-0.2in}
\end{figure*}

\section{Related Work}
We first review the relationship and difference between FSS-1000 and modern datasets aiming to solve image segmentation and few-shot classification. Then we review contemporary research on few-shot learning and semantic segmentation and discuss how we relate the few-shot segmentation to previous research.
\vspace{-0.1in}
\paragraph{Large-Scale Datasets} When deep learning had started to become a dominating tool for computer vision, the importance of building large-scale datasets was emphasized for training deep networks. The PASCAL VOC~\cite{voc} was the first to provide a challenging image dataset for object class recognition and semantic segmentation. The latest version VOC2012 contains 20 object classes and 9,993 images with segmentation annotations.
Despite the absence of segmentation labels, the Imagenet~\cite{imagenet} is built upon the backbone of WordNet and provides image-level labels for 5,247 classes for training, out of which a subset of 1,000 categories are split out to form the ILSVRC~\cite{ilsvrc} dataset. This challenge has made a significant impact on the rapid progress in visual recognition task and computer vision in recent years. The latest Open Image dataset~\cite{openimage} contains 7,186 trainable distinct object classes for classification and 600 classes for detection, making it the largest existing dataset with object classes and location annotations. Following the PASCAL VOC and ImageNet, the COCO segmentation dataset~\cite{coco} includes more than 200,000 images with instance-wise semantic segmentation labels. There are 80 object classes and over 1.5 million object instances in COCO dataset. \par
In this paper, we instead focus on broadening the number of object classes in a segmentation dataset rather than increasing dataset size. Our FSS-1000 consists of 1,000 object classes, wherein each class we label 10 images with binary segmentation annotation. So in total, our dataset contains 10,000 images with pixelwise segmentation labels. We are particularly interested in segmentation due to its obvious benefits: segmentation captures the essential feature of an object without background; instance level segmentation can be ready from segmentation. The structure of our dataset is similar to widely-used datasets for few-shot visual recognition. For example, the Omniglot dataset~\cite{omniglot} consists of 1,623 different handwritten characters of 50 different alphabets, which is equivalent to 1,623 object classes with 50 images in each class. The MiniImageNet, first proposed in~\cite{miniimagenet}, consists of 60,000 images with 100 classes each having 600 examples. But none of these few-shot learning datasets incorporate dense pixelwise segmentation labels, which is essential in training a deep network model for semantic segmentation.
\vspace{-0.1in}

\paragraph{Few-Shot Learning}  Recent research in few-shot classification can be classified into 1) learn a good initial condition for the network to be fine-tuned on extremely small training set, as proposed in~\cite{modelag,op}; 2) rely on memory properties of RNN, introduced in~\cite{metanetwork,memoryaug}; 3) learn a metric between few-shot samples and queries, as in~\cite{feedforward,prototype,omniglot,siamese,ltc}. We choose to extend the relation network~\cite{ltc} for few-shot segmentation because it is a simple, general and working framework. By concatenating the CNN feature maps between support images and query images, the relation module can consider the hidden relationship between these two sets of images guided by the loss function. In the original relation network, it uses the MSE loss to compare the final probability vector to the ground truth. In this paper, we simply modify the loss to calculate pixelwise differences between the segmentation ground truth and heatmap. In OSLSM~\cite{oslsm}, the authors proposed a two-branch network to solve few-shot segmentation. The network is quite complex, and their training set was limited to the PASCAL VOC dataset with only 20 object classes. Consequently, their feature extractor may suffer severe bias making it hard to be generalized to other objects. The guided network~\cite{iclr18} can also suffer the same limitation on their dataset choice. Though point annotation can be used to guide the training of few-shot segmentation, the sparse annotation can seriously hamper accuracy.

\vspace{-0.28in}

\paragraph{Semantic Image Segmentation}
Previous research exploiting CNN to make dense prediction often relied on patchwise training~\cite{seg1,seg2,seg3} and pre- and post-processing of superpixels~\cite{seg2,simul}. In~\cite{fcn} the authors first proposed a simple and elegant fully convolutional network (FCN) to solve semantic segmentation. Notably, this is the first work which was trained end-to-end on a fully convolutional network for dense pixel prediction, which showed that the last layer feature maps from a good backbone network such as VGG-16 contain sufficient foreground features which can be decoded by the upsampling network to produce segmentation results. Intuitively, that is also the guiding principle behind our modification on relation network architecture. Though modern network architectures~\cite{maskrcnn,mnc,fcis} achieve high accuracy in the COCO challenge by adding complex network modules and branches, these models cannot be adapted easily to segment new classes with few training examples.

\section{FSS-1000}
Recent few-shot datasets~\cite{omniglot, miniimagenet} support few-shot classification but there is no large-scale few-shot segmentation dataset.  In this section, we first introduce the details of data collection and annotation, then discuss the properties of FSS-1000. Table~\ref{tab:compare} and Figure~\ref{fig:compare} compare FSS-1000 with existing popular datasets. FSS-1000 targets at solving general objects few-shot segmentation problem. So datasets only focusing on sub-domain object categories in the world (e.g. handwritten characters, human faces and road scenes) are not included in the comparison.

\subsection{Data Collection}
\label{sec:data_collection}

\paragraph{Object Classes} We first referred to the classes in ILSVRC~\cite{ilsvrc} in our choice of object categories for FSS-1000. Consequently, FSS-1000 has 584 classes out of its 1,000 classes overlap with the classes in the ILSVRC dataset. We find ILSVRC dataset heavily biases toward animals, both in terms of the distribution of categories and number of images. Therefore, we fill in the other 486 by new classes unseen in any existing datasets. Specifically, we include more daily objects so that network models trained on FSS-1000 can learn from diverse artificial and man-made objects/features in addition to natural and organic objects/features where the latter was emphasized by existing large-scale datasets. Our diverse 1,000 object classes are further arranged in a hierarchy to be detailed in section~\ref{sec:properties}.

\vspace{-0.15in}
\paragraph{Raw Images} To avoid bias, the raw images were retrieved by querying object keywords on three different Internet search engines, namely, Google, Bing and Yahoo. We downloaded the first 100 results returned (or less if less than 100 images were returned) from a given search engine. No special criteria or assumption was used to select the candidates, however, due to the bias of Internet search engines, a large number of the images returned contain a single object photographed with sharp focus. In the final step, we intentionally included some images with a relatively small object, multiple objects or other objects in the background
to balance the easy and hard examples of the dataset.

Images with aspect ratio larger than 2 or smaller than 0.5 were excluded. Since all images and their segmentation maps were to be resized to $224 \times 224$, bad aspect ratio would destroy important geometric properties after the resize operation. For the same reason, images with height or width less than 224 pixels were discarded because they would trigger upsampling which would affect the image quality after resizing. 
\vspace{-0.15in}

\begin{figure*}[t]
\begin{center}
   \includegraphics[width=1.0\linewidth]{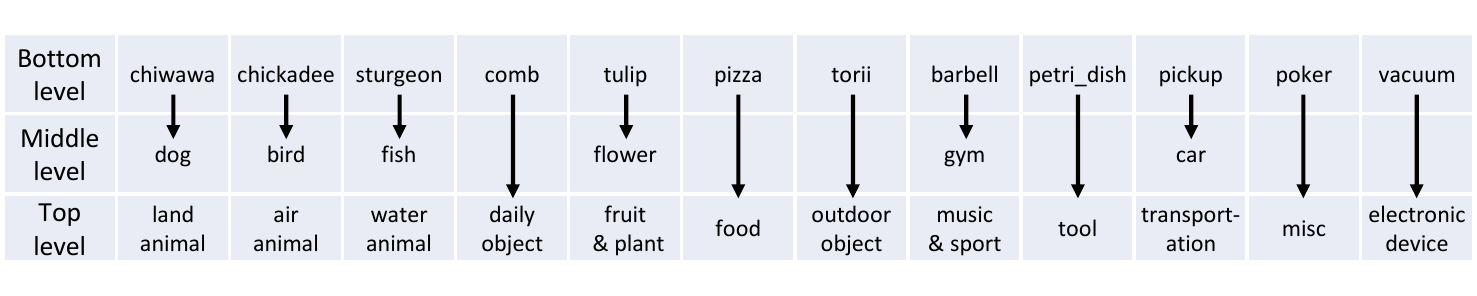}
\end{center}
\vspace{-0.3in}
   \caption{Hierarchy of FSS-1000. Arrow represents ``is a subclass of" relationship.}
\label{fig:hierarchy}
\vspace{-0.2in}
\end{figure*}

\begin{figure}[t]
\begin{center}
  \includegraphics[width=0.15\linewidth]{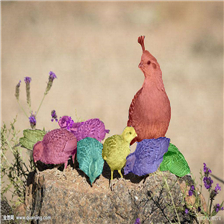}
  \includegraphics[width=0.15\linewidth]{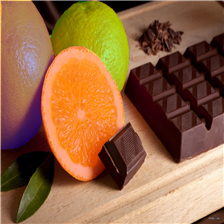}
  \includegraphics[width=0.15\linewidth]{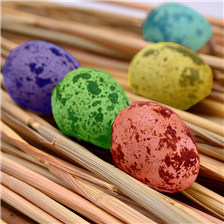}
  \includegraphics[width=0.15\linewidth]{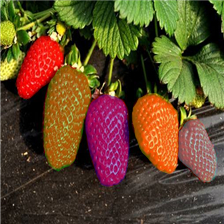}
  \includegraphics[width=0.15\linewidth]{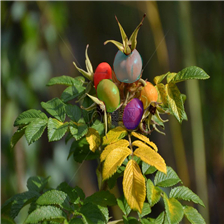}
  \includegraphics[width=0.15\linewidth]{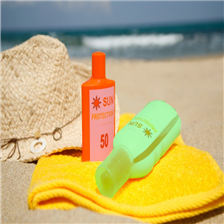}
\end{center}
\vspace{-0.1in}
  \caption{\red{Example of instance annotation in the FSS-1000 dataset.}}
\label{instance}
\vspace{-0.2in}
\end{figure}

\vspace{-0.1in}

\paragraph{Pixelwise Segmentation Annotation} We used Photoshop's ``quick selection" tool which allows users to loosely select an object automatically, and refined or corrected the selected area to produce the desired segmentation.
Figure~\ref{fig:examples} shows example images overlaid with their corresponding segmentation maps in FSS-1000.

\subsection{Properties}
\label{sec:properties}
This section summarizes the three desirable properties of FSS-1000:
\vspace{-0.1in}
\paragraph{Scalability}
To extend FSS-1000 to include a new class, all it takes are 10 images with pixelwise binary segmentation labels for the new class.
This is significantly easier than other datasets such as PASCAL VOC and COCO. First, the mean number of images in a given class is much larger than 10 in these datasets. Second, in these large-scale datasets the object classes need to be first pre-defined. 
Thus we believe binary annotation is a better annotation strategy in few-shot learning datasets, since it allows easy expansion of new object classes without concerning old object classes that have already been annotated.

\vspace{-0.1in}
\paragraph{Hierarchy}
Figure~\ref{fig:hierarchy} shows examples of one sub-category for each given super-category in the dataset to illustrate the hierarchical structure of FSS-1000. The object classes are arranged hierarchically following a 3-level structure, while not every bottom-level subclass has a middle-level superclass. The top of the object hierarchy consists of 12 super-categories while the bottom contains the 1,000 classes as the leaf nodes. Note that this is strictly not a tree structure because a given class may belong to more than one superclass (e.g., an apple is both ``fruit" and  ``food").

\vspace{-0.1in}
\paragraph{Instance}
FSS-1000 dataset supports instance-level segmentation with instance segmentation labels in 758 out of the 1,000 classes in the dataset, which are significantly more classes than PASCAL VOC and MS COCO. One major difference between our dataset and PASCAL VOC / MS COCO instance level segmentation is that our dataset only annotates one type of objects in one image, despite there may be other object categories appearing in the background. We annotate at most 10 instances in a single image, which follows the same instance annotation principle adopted by COCO. 
\red{Figure~\ref{instance} shows examples of instance annotations in the dataset.}

\begin{figure}[t]
\begin{center}
   \includegraphics[width=0.99\linewidth]{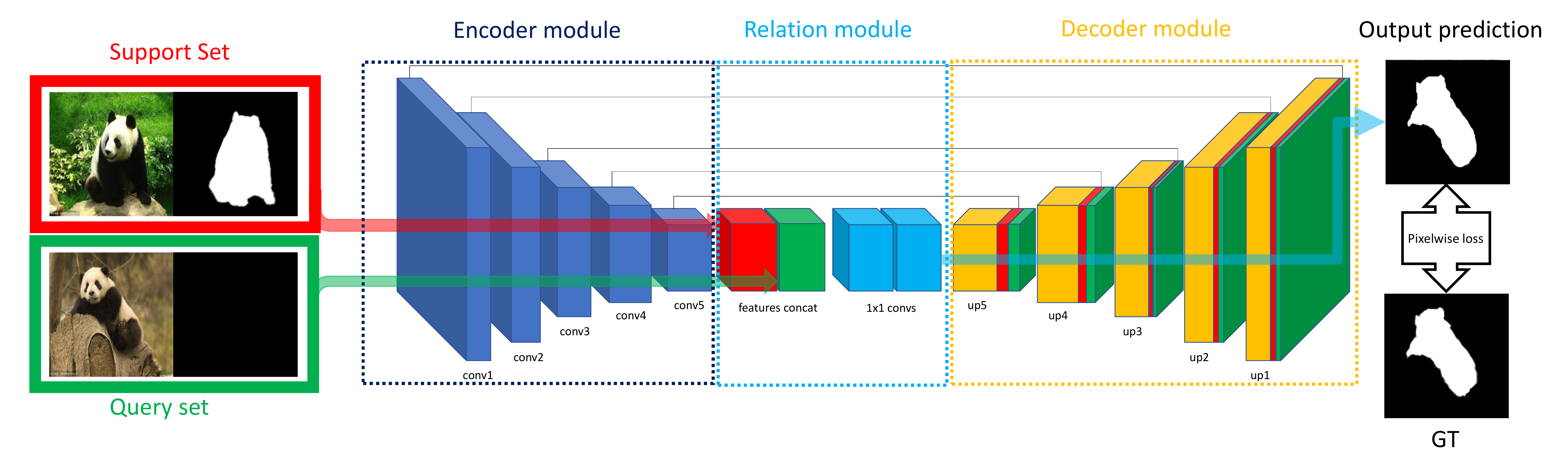}
\end{center}
\vspace{-0.2in}
   \caption{Our baseline network architecture using VGG-16 as backbone.  The relation module is adapted from~\cite{ltc} where a decoder module is appended to produce the segmentation map. Both support and query features are concatenated to the decoder module via skip connection. More details of this standard architecture are available in supplemental materials.}
\label{fig:network}
\vspace{-0.2in}
\end{figure}

\section{Methodology}
\subsection{Problem Formulation}
In few-shot learning, the train-test split is on \textit{object categories}, thus, all testing categories are unseen during training. In both training and testing, the input is divided into two sets, namely, the support set and the query set. The support set consists of samples with annotation, while the query set contains samples without annotation. In few-shot classification, the support set usually includes $C$ classes and $K$ training examples. This setting is defined as $C$-way-$K$-shot classification~\cite{feifei06oneshot, ltc}. In few-shot segmentation, we adopt this notation but extend the query output to be per-pixel classification of the query image, rather than a single class label. Specifically, in few-shot segmentation, the input-output pair is given by $(X,Y)$, where \\
$$L=\left\{l_{(i,j)};l\in\left\{1,2,...,C\right\}\right\} $$
$$X=\left\{(I_s,L_s, I_q); s\in\left\{1,2,...,K\right\}\right\}$$
$$Y=\left\{y_{(i,j)}|I_q; y\in\left\{1,2,...,C\right\}\right\}$$
$l_{(i,j)}$ is the ground-truth class label and $y_{(i,j)}$ represents the predicted class label for pixel $(i,j)$ in a given image. $I_s$ is the 3-channel RGB support image. For each support input $X$ with image and label pair $(I_s,L_s)$, the model predicts a pixelwise classification map over query image $I_q$. Following the annotation strategy of FSS-1000, we set $C=2$ and only focus on few-shot binary segmentation problem in this paper. However, a general $C$-way-$K$-shot segmentation could be solved by a union of $C$ binary segmentation tasks.

\subsection{Network Architecture}
\paragraph{Pipeline} Our network consists of three sub-modules: an encoder module $E_\theta$, a relation module $R_\phi$ and a decoder module $D_\omega$. For a given input $X$ to the network, the encoder $E_\theta$ encodes the support and query images respectively into feature maps $E_\theta(I_s)$ and $E_\theta(I_q)$.
For $K$-shot forwarding, we perform element-wise averaging over the depth channels of support feature maps, so that the encoder module always produces support feature maps of the same depth regardless of the size of the support set.

The support and query feature maps are then combined in the relation module $R_\phi$. We choose channel-wise concatenation as the combination operation, while other choices such as parameter regression and nearest neighbors are possible and discussed in~\cite{iclr18}. The relation module generates coarse segmentation results in low-resolution based on the concatenated feature maps. Finally, the coarse result is fed into the decoder module to restore the prediction map to the same resolution of the input. Figure~\ref{fig:network} shows the entire workflow. In summary, the output is defined by
$$Y=D_\omega(R_\phi(\sum_{s=1}^K E_\theta(I_s),E_\theta(I_q))).$$ 

\vspace{-0.25in}
\paragraph{Loss function} We use the cross entropy loss between the query prediction output and the ground-truth annotation to train our model. Specifically, under our binary few-shot segmentation setting, binary cross entropy (BCE) loss is adopted to optimize the parameters in the network:
\begin{align*}
\theta^*,\phi^*,\omega^* &=\\ \argmin\limits_{\theta,\phi,\omega}\sum_{i}\sum_{j}& -L_{(i,j)}\log y_{(i,j)}+(1-L_{(i,j)})\log(1-y_{(i,j)})
\end{align*}
Mean square error (MSE) is also a widely used objective function for semantic segmentation task. Different from BCE loss, MSE models the problem as regression to the target output. Our experiments show that BCE and MSE loss achieve similar performance under our network setting.

\subsection{Network Module Details}
One can design his/her own or choose any popular feature extraction backbone such as VGG-16~\cite{vgg}, ResNet~\cite{resnet} and Inception~\cite{inceptionnet} as the encoder module inside the network.
The support and query features compose the combined feature map whose depth is twice the channel number of the last-layer output of the encoder. The relation module utilizes two $1\times1$ convolutional layers on the combined feature map to embed the relationship between the support features and query features. The decoder module is designed according to the number of downscale operations in the encoder module, which applies equivalent upsample blocks to restore the resolution back to the original input. In each upsample block stands a nearest neighbor upsampling layer and a convolutional layer. Skip connection is adopted between encoder and decoder feature maps, following the scheme proposed by U-Net~\cite{unet}. We find it helpful to produce fine details in segmentation when information in the encoder feature maps are fused to the decoder module by channel-wise concatenation. ReLU activation is applied throughout the deep network except for the last layer's activation where Sigmoid is used in order to scale the output to a suitable range to calculate cross-entropy loss. More detail parameters of our architecture are provided in the supplemental materials.

\section{Experiments}

We conduct experiments to evaluate the practicability of FSS-1000 and the performance of our method under few-shot learning settings. We evaluate models with the same network architecture but trained on different datasets to show that FSS-1000 is effective for few-shot segmentation task. Different support sets and their influence on query results will be discussed. Finally we illustrate that models trained on FSS-1000 are capable to generalize the few-shot segmentation knowledge to new unseen classes.
The metric we use is the intersection-over-union (IoU) of positive labels in a binary segmentation map. IoU is a standard metric and widely adopted in evaluating image segmentation methods.
All the networks are implemented in PyTorch. We use Adam solver~\cite{adam} to optimize the parameters. The learning rate is initially set to $10^{-3}$ ($10^{-4}$ for fine-tuning) and halved for every $50,000$ episodes. We train all the networks for $500,000$ episodes.\par

\begin{table}[t]
\begin{center}
\begin{adjustbox}{max width=0.5\linewidth}
\begin{tabular}{l|c}
\hline
Method & MeanIoU\\
\hline
\textbf{VGG-16-BCEloss} & \textbf{80.12\%}\\
VGG-16-MSEloss & 79.66\%\\
ResNet-101-BCEloss & 79.43\%\\
ResNet-101-MSEloss & 79.12\%\\
InceptionV3-BCEloss & 79.02\%\\
InceptionV3-MSEloss & 79.22\%\\
\hline
\end{tabular}
\end{adjustbox}
\end{center}
\vspace{-0.1in}
\caption{Different network settings to explore the best settings for our network architecture.}
\vspace{-0.15in}
\label{networksetting}
\end{table}


\noindent {\bf Network setting~} To explore the best settings for our network, we train different models using a combination of different backbones and loss functions on FSS-1000. Table~\ref{networksetting} tabulates the respective performance on VGG-16, ResNet-101 and InceptionNet as backbone, and BCE and MSE as loss function. Based on the result, we choose VGG-16 as feature extractor and use BCE loss in our model throughout the experimental section.

\begin{table}[t]
\begin{center}
\begin{adjustbox}{max width=0.6\linewidth}
\begin{tabular}{l|c}
\hline
Method & MeanIoU\\
\hline
OSLSM-1shot~\cite{oslsm}& 70.29\% \\
OSLSM-5shot & 73.02\%\\
Guided Network-1shot~\cite{iclr18}& 71.94\%\\
Guided Network-5shot & 74.27\%\\
Ours-1shot & 73.47\%\\
\textbf{Ours-5shot} & \textbf{80.12}\%\\
\hline
\end{tabular}
\end{adjustbox}
\end{center}
\vspace{-0.1in}
\caption{Different few-shot segmentation networks trained and tested on FSS-1000.}
\label{fewshotmethods}
\vspace{-0.15in}
\end{table}

\begin{table}[t]
\begin{center}
\begin{adjustbox}{max width=0.99\linewidth}
\begin{tabular}{l|c|c|c|c|c}
\hline
Method & PASCAL-$5^0$ & PASCAL-$5^1$ & PASCAL-$5^2$ & PASCAL-$5^3$ & Mean\\
\hline
OSLSM~\cite{oslsm} & 34.2\% & 57.9\% & 43.2\% & 37.8\% & 43.3\%\\
GN~\cite{iclr18} & 33.1\% & 58.9\% & 44.3\% & 39.9\% & 44.1\%\\
Ours & 37.4\% & 60.9\% & 46.6\% & 42.2\% & 46.8\%\\
\red{PANet}~\cite{panet} & 51.8\% & 64.6\% & \textbf{59.8\%} & 46.5\% & 55.7\% \\
\red{CANet}~\cite{canet} & \textbf{55.5\%} & 67.8\% & 51.9\% & 53.2\% & 57.1\% \\
Ours* & 50.6\% & \textbf{70.3\%} & 58.4\% & \textbf{55.1\%} & \textbf{58.6\%}\\
\hline
\end{tabular}
\end{adjustbox}
\end{center}
\vspace{-0.1in}
\caption{Comparison of different models on PASCAL-$5^i$. GN is Guided Network and Ours* is our model trained on FSS-1000. All models are using 5-shot setting.}
\vspace{-0.25in}
\label{pascal5i}
\end{table}



\subsection{Benchmarks}
\subsubsection{FSS-1000}
We train OSLSM and Guided Network on FSS-1000 to provide benchmarks and justify our dataset. Table~\ref{fewshotmethods} shows that our adapted relation network achieves the best results on FSS-1000. Moreover, ours is the only model whose 5-shot training boosts the accuracy by over 10\% compared to the 1-shot case. We believe that embedding multiple support images at the input end of the network and encouraging the feature extractor to consider correlation between multiple support images and the query image is the appropriate way to design $k$-shot ($k>1$) segmentation network, rather than simply combining 1-shot prediction~\cite{oslsm} or merging high-level features of multiple supports~\cite{iclr18}.

\subsubsection{PASCAL-$5^i$} To compare with previous few-shot methods, we train and test our network on PASCAL-$5^i$~\cite{oslsm}. 
Table~\ref{pascal5i} shows that our simple baseline model (Ours) marginally outperforms OSLSM and Guided Network. \red{More importantly, our model trained only on FSS-1000 without fine-tuning on PASCAL-$5^i$ (Ours*) achieves much better results compared to models trained on PASCAL-$5^i$ (Ours), exceeding the state-of-the-art performance of the very recent~\cite{canet,panet}.}

\begin{table}
\begin{center}
\begin{adjustbox}{max width=0.9\linewidth}
\begin{tabular}{c|ccc|c|c}
\hline
No. & ImageNet & FSS & fsCOCO  & FSS (test set) & fsCOCO (test set)\\
\hline
\uppercase\expandafter{\romannumeral1}&\cmark &  & \cmark &  71.34\% & 42.11\%\\
\uppercase\expandafter{\romannumeral2}&\cmark &\cmark  & &  79.30\% & 47.99\%\\
\uppercase\expandafter{\romannumeral3}& &\cmark  & &  80.12\% & 48.31\%\\
\uppercase\expandafter{\romannumeral4}& &\cmark &\cmark & 82.66\% & 50.56\%\\
\hline
\end{tabular}
\end{adjustbox}
\end{center}
\caption{\blue{Comparison of models trained and tested on different datasets. Each model (row) shows the training stages, e.g., model~I uses the pre-trained weights from ImageNet then fine-tuned on fsCOCO's training classes, and finally tested on the novel test classes in both FSS and fsCOCO. All learning rates are initially set to $10^{-4}$ except the model trained without using ImageNet pre-trained weights, which is set to $10^{-3}$.}}
\label{diffdatasets}
\vspace{-0.1in}
\end{table}

\begin{figure}[t]
\begin{center}
   \includegraphics[width=0.9\linewidth]{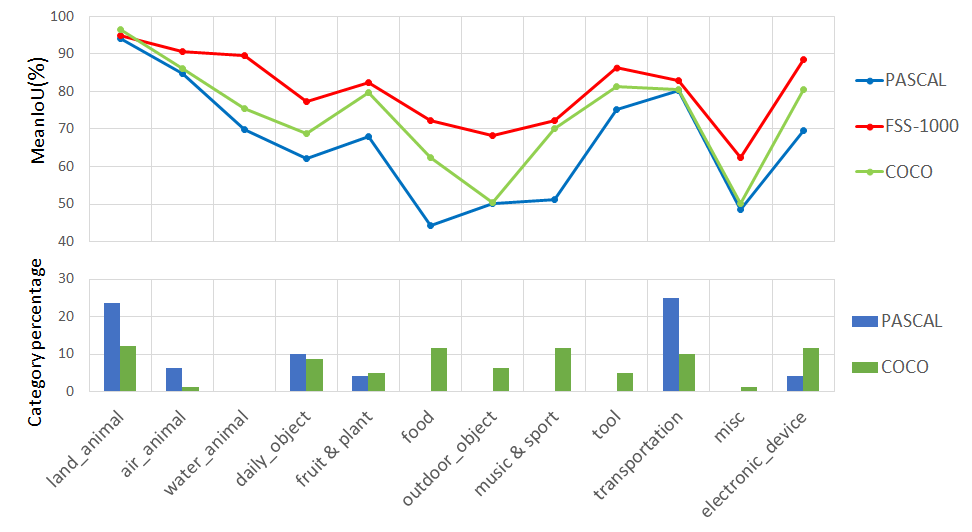}
\end{center}
\vspace{-0.2in}
   \caption{MeanIoU of superclasses in FSS-1000 tested with models trained on fsPASCAL, fsCOCO and FSS-1000. Bars at the bottom indicate the percentage of the number of categories overlapping with FSS-1000 in the corresponding dataset.}
\label{distribution}
\vspace{-0.25in}
\end{figure}

\begin{figure}[t]
\begin{center}
   \includegraphics[width=0.99\linewidth]{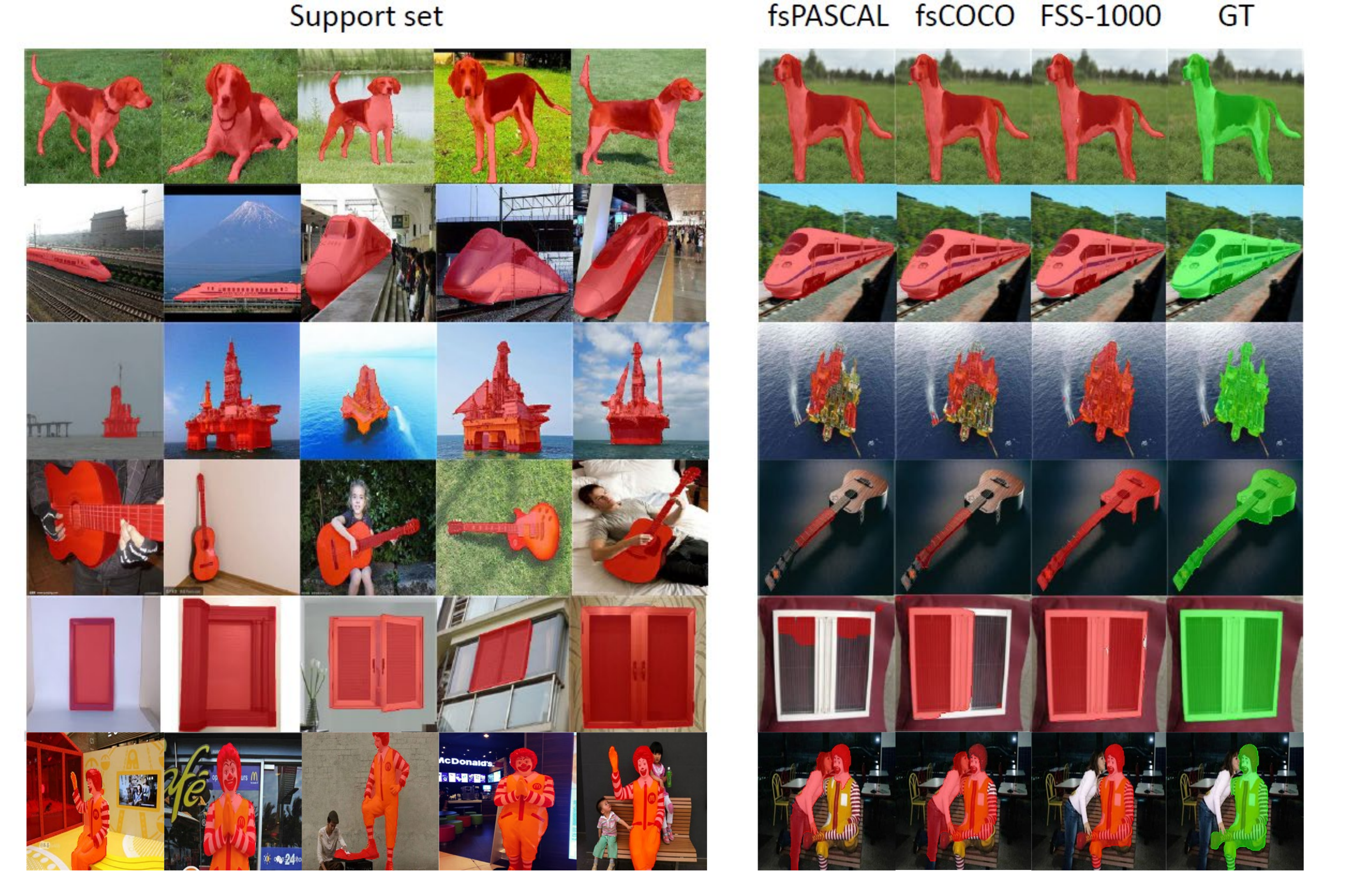}
\end{center}
\vspace{-0.2in}
   \caption{Image results of our baseline model respectively trained on
fsPASCAL, fsCOCO and FSS-1000. Support labels and predicted segmentation are
overlaid in red in corresponding support images and query images. Ground
truth labels for query images are in green. The classes
in the first two rows are present in fsPASCAL and fsCOCO whereas the rest are unique in FSS-1000.}
\label{imagecompare}
\vspace{-0.1in}
\end{figure}

\begin{figure}[t]
\begin{center}
   \includegraphics[width=0.9\linewidth]{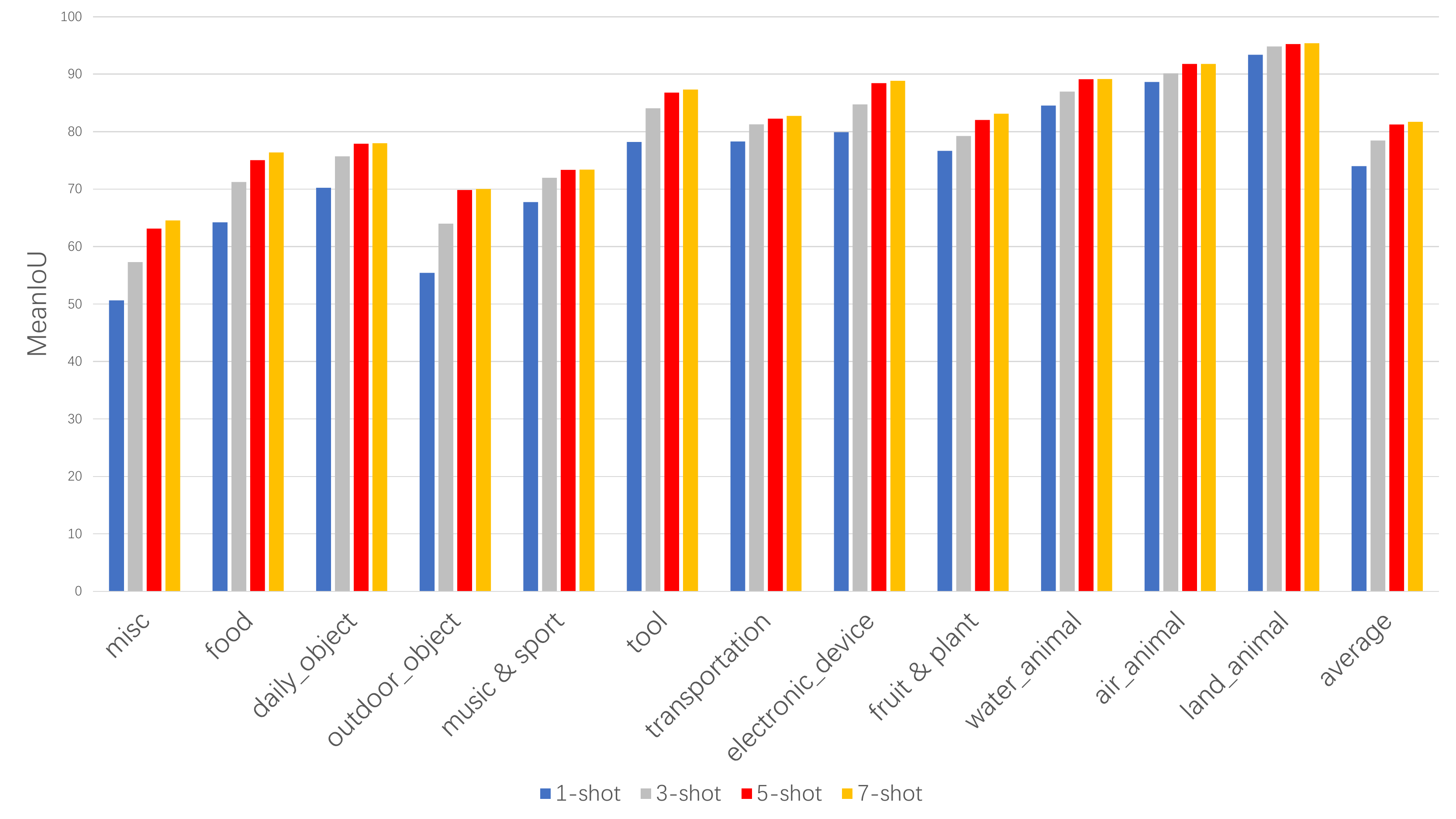}
\end{center}
\vspace{-0.2in}
   \caption{MeanIoU of superclasses in FSS-1000 tested with k-shot models (k = 1,3,5,7).}
\label{fig:kshot}
\vspace{-0.1in}
\end{figure}

\subsection {Effect of Pre-training} 
We compare our network model trained on different datasets to demonstrate the effectiveness of FSS-1000 in few-shot segmentation. Since there are no publicly available few-shot image segmentation datasets, we convert PASCAL VOC 2012 and COCO datasets by setting the desired foreground class label as positive and all others as negative, followed by the identical clean-up stage described in section~\ref{sec:data_collection} to the binarized labels. Two new datasets are thus produced: fsPASCAL and fsCOCO. There are respectively 4,318 image and label pairs in 20 object classes in fsPASCAL, \blue{which consists of 15 training classes and 5 test classes,} and 48,015 image and label pairs in 80 object classes in fsCOCO, \blue{containing 60 training classes and 20 test classes. The generation of these datasets are in line with the settings in \cite{canet}.}

For FSS-1000, we build the validation/test set by randomly sampling 20 distinct sub-categories from the 12 super-categories; the other images and labels are used in training.  The train/validation/test split used in the experiments consists of 5,200/2,400/2,400 image and label pairs. \blue{Each test set of fsPASCAL, fsCOCO and FSS are designed to be disjoint with all the training sets in terms of classes for fair comparison.}

Table~\ref{diffdatasets} tabulates the performance of different models. For each model (row), the \cmark marks in sequence indicate the dataset(s) used in pre-training stages with the last mark indicating the dataset used in fine-tuning. 
Model~\uppercase\expandafter{\romannumeral3} has only one \cmark indicating that it is exclusively trained on the dataset.

Using the pre-trained weights from ImageNet, Model~\uppercase\expandafter{\romannumeral2} trained on FSS-1000 outperforms the fsCOCO-trained model \uppercase\expandafter{\romannumeral1} \blue{on both test sets by a large margin of 8\% and 5.8\%, which is due to the FSS training set containing the COCO training classes, but with more variety.}
Notably, {\em without} using any pre-trained weights
Model~\uppercase\expandafter{\romannumeral3}  achieves slightly better results compared to Model~\uppercase\expandafter{\romannumeral2}, which substantiate our claim that bias in feature extractor does exist in models pre-trained and/or trained on a dataset unevenly distributed in object categories and images within each class.

\begin{figure}[t]
\begin{center}
   \includegraphics[width=0.99\linewidth]{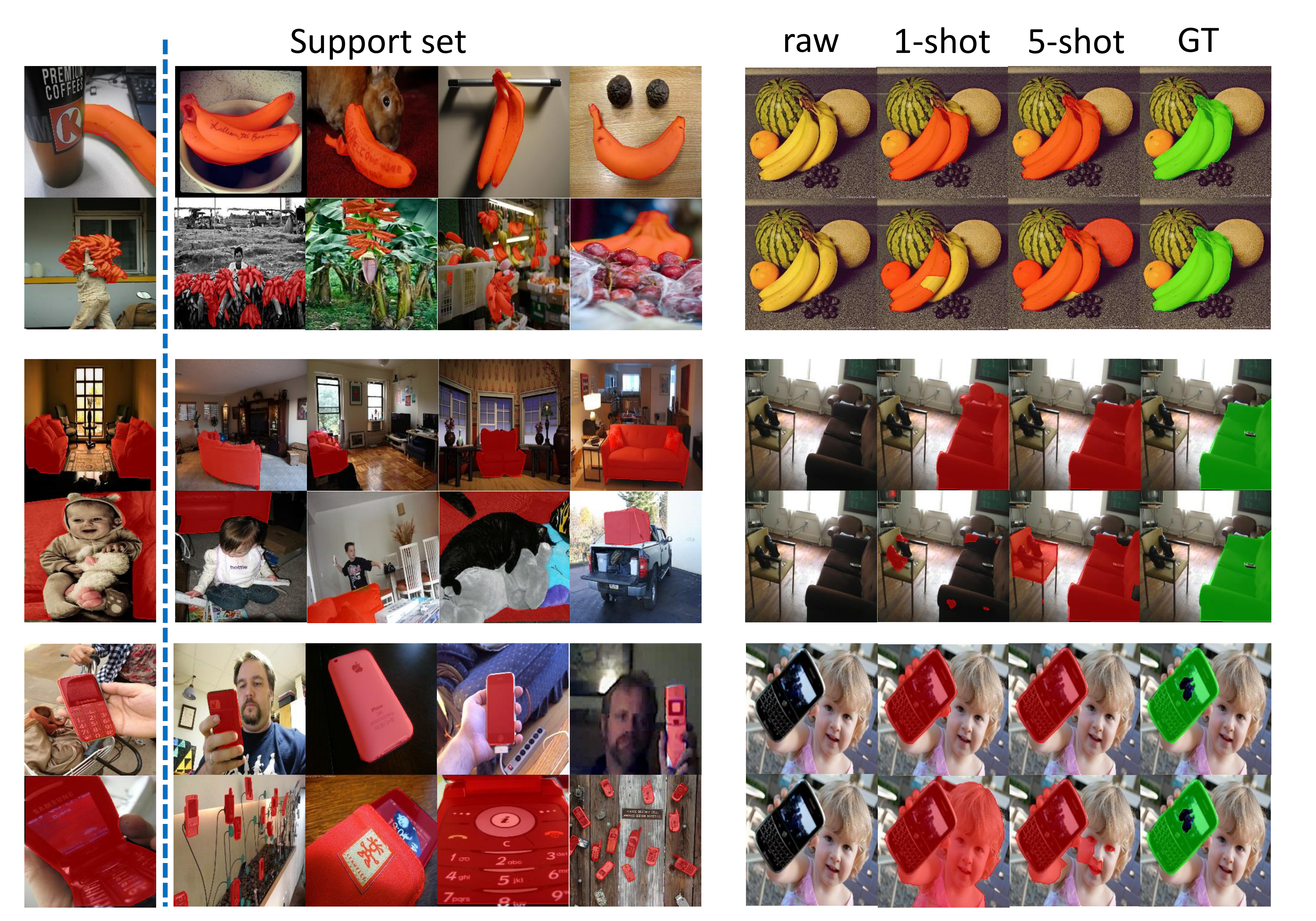}
\end{center}
\vspace{-0.15in}
   \caption{Effect of different support sets. The leftmost support of each row is used to generate 1-shot results. For each class, we show the result of a good support set followed by a bad support set in the next row.}
\label{supporteffect}
\vspace{-0.1in}
\end{figure}

\begin{table}
\begin{adjustbox}{max width=\columnwidth}
\begin{tabular}{ccccc}
                 & Human (PS) & Human (GrabCut)  & CPU   & GPU   \\
\hline
Time    & 180m32s & 53m22s & 9m13s & \textbf{16.9s} \\
95\%+ IOU  & 100\% & 71.4\%    & 58.4\%   & 58.4\%   \\
90\%+ IOU  & 100\% & 80.4\%   & 70.4\%   & 70.4\%   \\
80\%+ IOU  & 100\% & 91.0\%   & 87.4\%   & 87.4\%   \\
70\%+ IOU  & 100\% & 95.8\%   & 90.2\%   & 90.2\% \\ \hline
\end{tabular}
\end{adjustbox}
\caption {500 test images are randomly sampled from FSS-1000 to compare time and accuracy performance of labeling segmentation data between humans and few-shot model.}
\label{table:time}
\vspace{-0.2in}
\end{table}

Interestingly, \blue{Model~\uppercase\expandafter{\romannumeral4} pre-trained on FSS-1000 and fine-tuned on fsCOCO achieves the best result on both test sets}, outperforming \blue{Model~\uppercase\expandafter{\romannumeral3} exclusively trained on FSS and} the model~\uppercase\expandafter{\romannumeral1} pre-trained on ILSRVC fine-tuned on fsCOCO. We believe \blue{the former is due to the addition of more data, and the latter} is due to the difference in requirement of feature maps ideal for classification and segmentation task. Intuitively, semantic segmentation requires more accurate low-level features to produce fine details in segmentation map, while classification focuses on high-level features for image understanding. \blue{Therefore, we argue that pre-training with FSS-1000 serves as a good alternative for ImageNet pre-training in few-shot semantic segmentation.}

Overall, models trained on fsCOCO produce quite good results in \blue{test classes that are similar to COCO training classes.} For these classes, sometimes their segmentation results are better in local details compared to the results produced by models trained on FSS-1000 due to more variations in the training set. However, it failed in \blue{classes significantly different from the 60 COCO training classes}. The somewhat limited variation in object categories in existing datasets makes it hard for models trained on them to generalize to more unseen classes under the few-shot setting.
On the other hand, models trained on FSS-1000 classes can handle these cases. Quantitative results and qualitative results are shown in Figure~\ref{distribution} and Figure~\ref{imagecompare} respectively. Results on fsPASCAL and further comparisons are provided in supplementary material.

\subsection{Effect of Support Set}
\label{suppeffect}
We train four different models, using 1, 3, 5 and 7 support images respectively, to study how different number of support images influence the accuracy of few-shot segmentation. Two important observations can be summarized from Figure~\ref{fig:kshot}.

First, more support images generally boost the segmentation accuracy because more variations of color, pose, and scale of the object are included. However, the performance increase becomes negligible when more than 5 support images are given. Due to this bottleneck effect, we set up most of the experiments under the 5-shot setting.

Second, the accuracy boost is different among different classes. For easy cases (e.g. rigid objects), the improvement is not obvious because a single support image is enough for the deep network to capture and distinguish strong features of the object. For hard cases (e.g. deformable objects), more support images are essential for the network to learn the complex shapes to make correct segmentation.

Figure~\ref{supporteffect} demonstrates the effect of support set, which shows that scale and pose of the object to be segmented are the most important characteristics to guide few-shot semantic segmentation on FSS-1000. Since FSS-1000 does not explicitly consider scale variations (future work), a tiny or oversized object in the support set is not a good reference for segmentation. Significant differences in scales can mislead the network to capture wrong feature contents in the query. Besides, significantly different poses in support and query sets can result in bad segmentation results, due to the intrinsic fragility to rotation in CNN features.

\subsection{Auto-Labeling on Novel and Unseen Classes}
Traditionally a large number of human-annotated images are required to train a deep network for segmenting a new class.  Table~\ref{table:time} tabulates the tradeoff in time and accuracy for annotating 500 test images in FSS-1000 by humans (using Photoshop and GrabCut~\cite{grabcut} algorithm) and our few-shot segmentation.

\begin{figure}[t]
\begin{center}
  \includegraphics[width=0.99\linewidth]{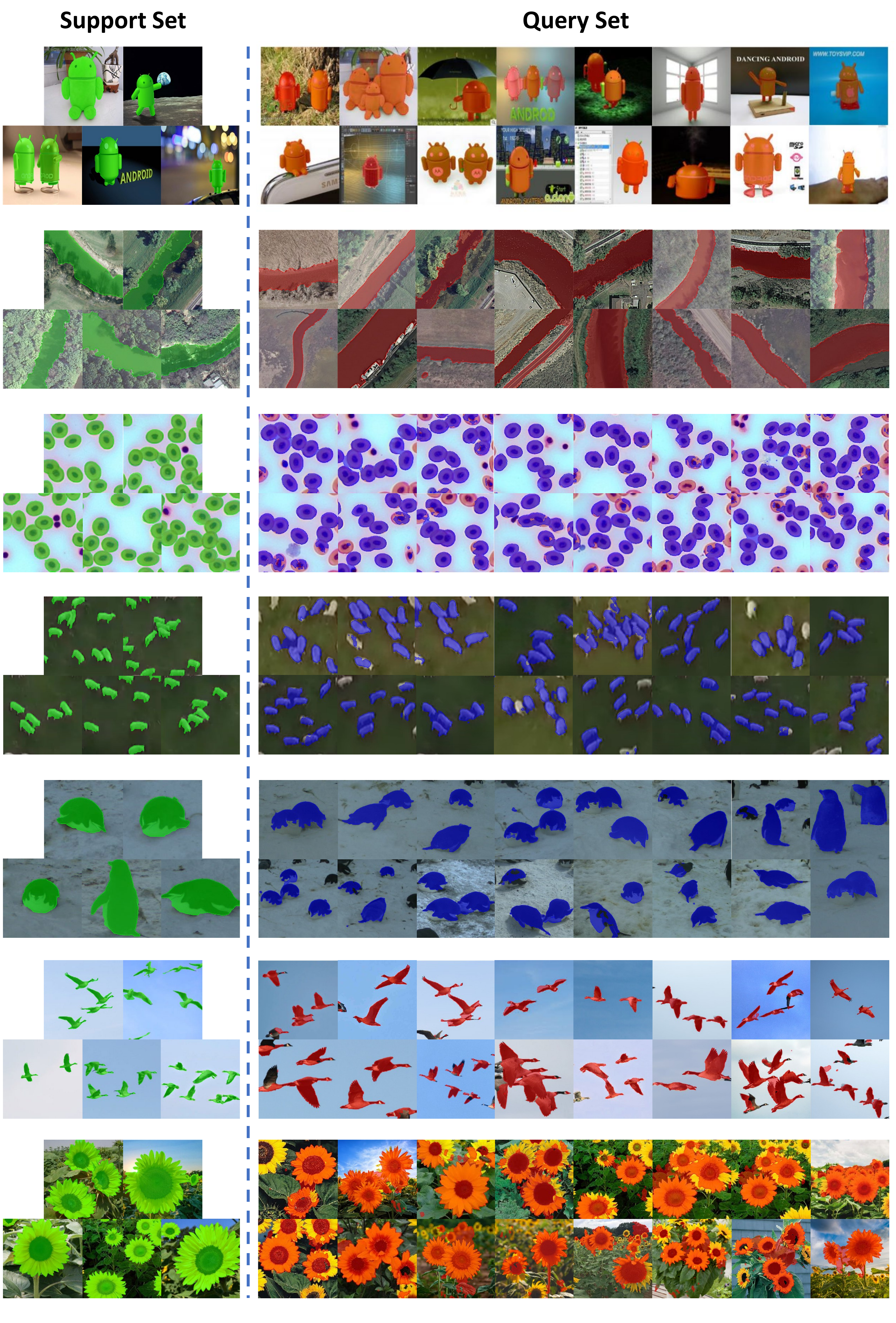}
\end{center}
\vspace{-0.2in}
  \caption{Test results for unseen classes. From top to bottom: {\em
android robot}; the {\em river} from UC Merced Land Use
Dataset~\cite{ucdata}; a large {\em cell} image cropped into patches;
herds of {\em sheep}; {\em penguin} from Oxford penguin counting
dataset~\cite{Arteta16}; flock of wild {\em goose}; different images of
fields of {\em sunflower} depict  various scales in the presence of
occlusion and perspective distortion. }
\vspace{-0.2in}
\label{fig:unseen}
\end{figure}

With its good accuracy and time tradeoff, despite the current limitations in scale invariance aforementioned,  FSS-1000 allows us to automatically segment a novel object category by just providing a few support examples {\em without} re-training or fine-tuning a given model.
We pick a number of very novel classes unseen by FSS-1000, and label 5 images of each class serving as the support set. Figure~\ref{fig:unseen} shows the test results which demonstrates that our model trained on FSS-1000 is capable of generalizing to these unseen classes. More extensive results on novel classes are included in supplementary materials.

For example, {\em android robot}  is an unreal object unseen in
FSS-1000.  In cartography from satellite images which often come in
overlapping image tiles, cartographers need to label only 5 images or
tiles and our system can automatically segment the rest, such as
recognizing {\em river} in our example where saliency detection does not
work in general.
The {\em cell} example shows the good potential of FSS-1000 in instance
segmentation which significantly contributes to cell counting in medical
image analysis where, for instance, a patient's health directly correlates
to his or her red blood cell count. With the advance of whole slide images
(WSI) in which the width and height often exceed 100,000 pixels (and thus
many cells to count), using our few-shot segmentation trained on FSS-1000,
pathologists only need to label 5 image relevant regions and then the rest
of the WSI will be automatically labeled. Although manual corrections for
missed or wrong cells may still be necessary given the current accuracy,
comparing with exhaustive labeling which requires hours or even days to
complete, the potential contribution of FSS-1000 is substantial. Similarly,
the related animal examples of {\em sheep}, {\em penguin} and wild {\em
goose} show FSS-1000's potential for large-scale instance segmentation.
Finally, our baseline backbone network is not very robust to
scale variance, occlusion and background noises (future work). In {\em
sunflower}, the segmentation results for instances too big or too small
(especially for images with depth of field where faraway sunflowers are
out of focus) become incomplete or even totally omitted. Despite that, FSS-1000
still reports limited success.

\subsection{Iterative Few-Shot Segmentation}

\begin{figure}[t]
\begin{center}
   \includegraphics[width=0.90\linewidth]{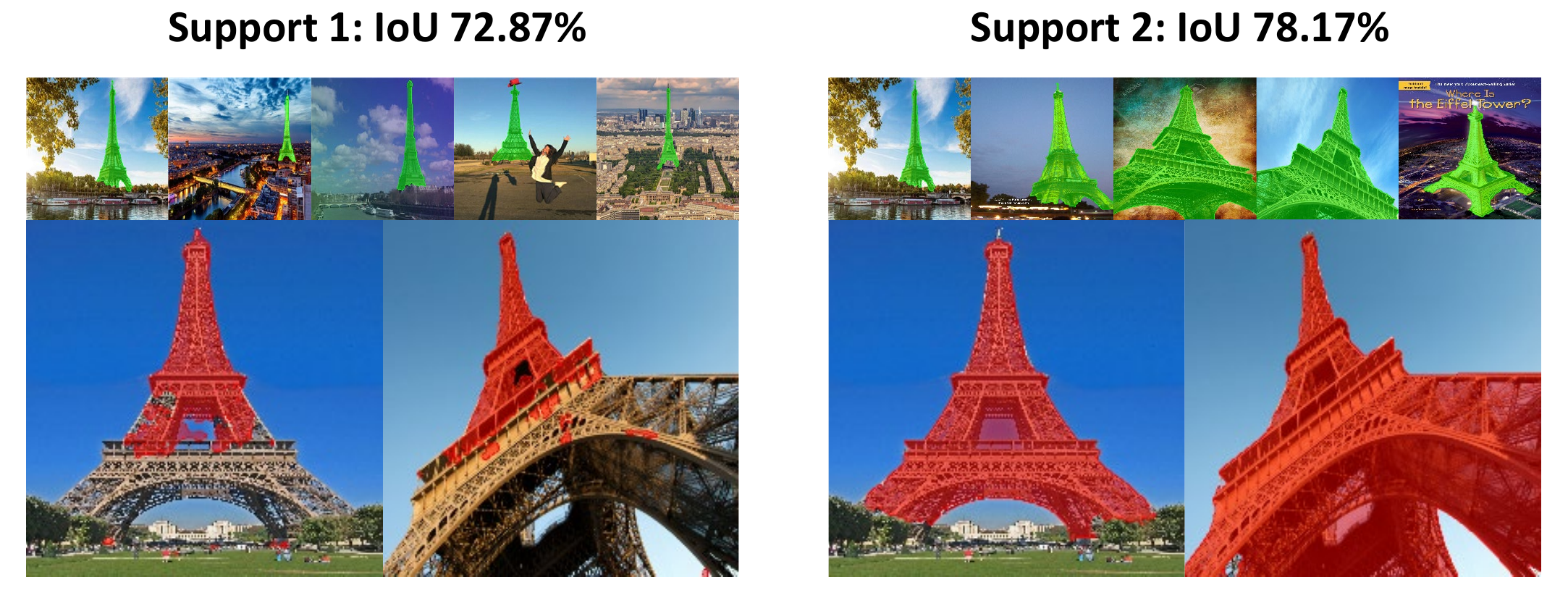}
\end{center}
\vspace{-0.2in}
\caption{Iterative few-shot segmentation. Left and right show respectively the support sets and results before and after including corrected failure cases in the support set. Complete testing set of {\em Eiffel Tower} is available in the supplemental material.}
\label{fig:hardcase}
\vspace{-0.1in}
\end{figure}

Our few-shot segmentation successively benefits from support sets improved easily by including failure cases after correction in each pass. Consider the Eiffel Tower unseen by FSS-1000 in Figure~\ref{fig:hardcase} where we manually label 200 images for quantitative evaluation (IoU). The first support set (left) did not have sufficient view and scale variations and did not see clearly the bottom part of the tower which resulted in its incomplete segmentation in some test cases. After mining a few of such hard cases, correcting and including them in the second support set (right),  the previous hard cases could now be correctly segmented. We believe that few-shot segmentation performed in stages  can offer an immediate performance boost.

\section{Conclusion}

Few-shot learning/segmentation is an emerging attractive alternative where only a few training examples are required. However, there is no existing large-scale dataset for few-shot segmentation. In this paper, we address the limitation of existing large-scale datasets in their biases and lack of scalability, and build the first few-shot segmentation dataset FSS-1000 emphasizing class diversity rather than dataset size. We adapt the relation network architecture to few-shot segmentation. This baseline few-shot segmentation model, trained exclusively on FSS-1000 without using pre-trained weights, achieves higher accuracy than previous methods including on test sets unseen by FSS-1000. We further demonstrated the efficacy and potential of FSS-1000 in large-scale segmentation on totally unseen classes without re-training or fine-tuning, and showed its promise on few-shot instance segmentation and iterative few-shot recognition tasks.

\clearpage

{\small
\bibliographystyle{ieee_fullname}
\bibliography{egbib}
}

\end{document}